\documentclass[conference]{IEEEtran}
\IEEEoverridecommandlockouts
\usepackage[numbers]{natbib}
\usepackage{graphicx}
\usepackage[hyphens]{url}
\usepackage{algorithm}
\usepackage{algorithmic}
\usepackage{booktabs}
\usepackage{amsmath}
\usepackage{amssymb}

\usepackage{amsthm}

\newtheorem{proposition}{Proposition}

\newcommand{\ind}{\mathbb{1}}

\begin{document}

\title{Who Makes the Decision? Auditing an Agentic LLM Pipeline for Illicit Cryptocurrency Transaction Monitoring}

\author{%
\IEEEauthorblockN{Vaishnavi Chaughule\IEEEauthorrefmark{1},
Yuchen Zhang\IEEEauthorrefmark{1},
Mohammed Ahnaf Tajwar\IEEEauthorrefmark{1},
Sai Akshay Menta\IEEEauthorrefmark{1},\\
Saanidhya Vats\IEEEauthorrefmark{1},
Vinija Jain\IEEEauthorrefmark{2},
Aman Chadha\IEEEauthorrefmark{3},
Divya Chaudhary\IEEEauthorrefmark{1}}
\IEEEauthorblockA{\IEEEauthorrefmark{1}Khoury College of Computer Sciences, Northeastern University, USA\\
\IEEEauthorrefmark{2}Meta, USA\\
\IEEEauthorrefmark{3}Apple, USA}}

\maketitle
\begingroup\renewcommand\thefootnote{}\footnotetext{\footnotesize Accepted to the 3rd Workshop on Big Data Analysis AND Illicit Trends (BANDIT) at IEEE BigData 2026. \copyright\ 2026 IEEE. Personal use of this material is permitted. Permission from IEEE must be obtained for all other uses, in any current or future media, including reprinting/republishing this material for advertising or promotional purposes, creating new collective works, for resale or redistribution to servers or lists, or reuse of any copyrighted component of this work in other works.}\addtocounter{footnote}{-1}\endgroup

\begin{abstract}
Agentic pipelines that combine graph detectors, deterministic rules, retrieval-augmented generation and an LLM judge are being proposed to monitor illicit cryptocurrency flows and to write regulation-citing justifications for analysts. They are evaluated on end-to-end accuracy or on the quality of those justifications, and neither shows which component made a decision. We audit such a pipeline on the Elliptic++ Bitcoin dataset with decision-source attribution, a trace-level protocol with three layers: per-case actual causation, interventional Shapley credit, and mediation of upstream risk scores. A deterministic wallet-score override carries about 94\% of the credit for flags, and a verification gate cancels most of the judge's own flags, which fall on licit transactions. Testing the judge alone over 76 evidence bundles, four backbones and two prompts, we find that it follows an injected risk score on the anonymous transaction features it is given, although stronger models decide correctly when the facts are explicit; that a context too short for the system prompt leaves it with no verdict at all; and that it cites regulatory clauses that do not apply, which a substring-based citation check accepts. Hiding the score does not fix the judge, because evidence gathered with the score visible still encodes it. For illicit-activity monitoring, an LLM's justification should therefore not be read as evidence of which component flagged a transaction.
\end{abstract}


\section{Introduction}

Cryptocurrency networks carry illicit flows from darknet-market sales, ransomware payments and laundering through mixers, and exchanges and investigators rely on graph-based detectors to find them \citep{weber2019antimoney,elmougy2023demystifying}. Financial crime compliance is moving from single detection models to pipelines of cooperating agents \citep{axelsen2025agentic,naik2025coinvestigator}. These systems are often called explainable because an LLM writes a regulation-citing justification for each case \citep{khanvilkar2025regulatory}, but whether that justification is reliable, or reflects the component that decided, is rarely tested. A justification is not the same as knowing who decided. When a graph model, a rule, a retriever, a verifier, and an LLM judge all touch a case, the component that determines the outcome can differ from the component that explains it.

This is an accountability problem. An auditor reviewing an escalation needs to attribute it to a responsible component, and that attribution has to survive delegation from one agent to the next. End-to-end accuracy cannot provide it: a pipeline can score well while its judge has no causal role. Explanation-quality ratings cannot provide it either, because a fluent justification can be written for a decision someone else made.

\begin{figure}[!t]
\centering
\includegraphics[width=\columnwidth]{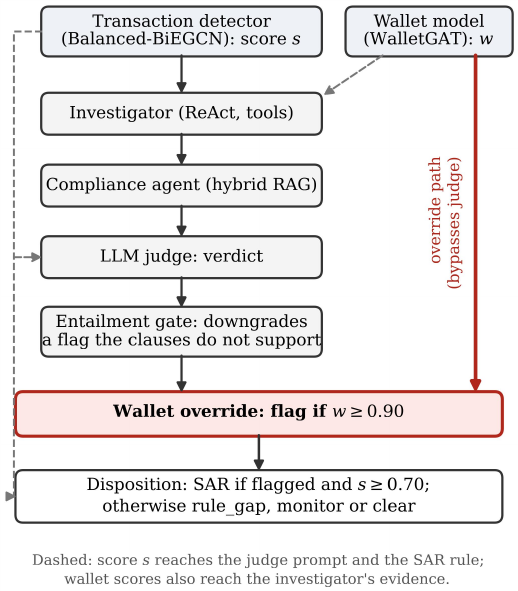}
\caption{The audited pipeline. The LLM chain (investigator, compliance agent, judge, gate) produces a verdict, but a deterministic wallet-score override can flag a case on its own (red path), without passing through the judge. The transaction score $s$ is shown to the judge and also decides escalation to a SAR.}
\label{fig:pipeline}
\end{figure}

We study this in a multi-agent AML pipeline on the Elliptic++ Bitcoin dataset \citep{elmougy2023demystifying}. An investigator agent gathers evidence with bounded tool use, a compliance agent retrieves regulatory clauses, an LLM judge issues a verdict, an entailment gate checks that verdict against the cited clauses, and a deterministic wallet-score override can flag a case on its own (Fig.~\ref{fig:pipeline}).

We ask three research questions. \textbf{RQ1:} which component decides the outcome of a case? \textbf{RQ2:} does the LLM judge decide from the evidence or from the upstream risk score? \textbf{RQ3:} do the answers hold under another pipeline structure, judge backbone, and out-of-sample cases, and do the detector-level findings transfer to a second dataset?

\paragraph{Why not an ablation}
Removing a component and measuring the change in accuracy answers a different question. An ablation reports the average effect of a component's absence. It does not say, for a given case, which component was an actual cause of that case's decision. It cannot assign signed credit when one component cancels another, and it cannot separate a judge that agrees with an override from a judge that copies the override's input. Decision-source attribution works on individual traces, asks counterfactual questions about each component's output, and states which conclusions follow from the design alone.

\paragraph{Contributions, in order of importance}
\begin{enumerate}
\item \textbf{An audit of an agentic AML pipeline} that answers the three questions. RQ1: once a truncation failure that made the judge look inert is fixed, a wallet-score override still carries about 94\% of flag credit. RQ2: on anonymous features the judge follows an injected score across 76 evidence bundles and four backbones and cites clauses that do not apply. RQ3: structure matters more than the backbone, and out of sample the pipeline does not separate illicit from licit cases; on AMLworld a label-free proxy override again carries the detection, by a thin margin, and the judge again follows the score.
\item \textbf{Decision-source attribution}, a protocol with three layers (per-case actual cause and responsibility, interventional Shapley credit, mediation of upstream signals), explicit null policies, matched negative controls with stated falsification conditions, and estimators validated against hand-derived answers. It needs only execution traces and the ability to intervene on component outputs.
\item \textbf{Structural propositions} that identify which attribution results carry no empirical content in a given pipeline, so that design consequences are not reported as findings.
\end{enumerate}

\section{Related Work}

\paragraph{Agentic financial crime compliance}
\citet{axelsen2025agentic} design an agentic system covering onboarding, monitoring, investigation, and reporting, evaluated formatively through stakeholder walkthroughs. \citet{naik2025coinvestigator} draft suspicious activity report narratives after activity is already flagged. \citet{khanvilkar2025regulatory} combine a GNN detector with retrieval-augmented GPT-4 explanations and rate explanation quality with experts. None of these measures which component causes each decision.

\paragraph{Two published pipelines}
We applied the protocol's interventions to two published pipelines with public code. OCR-APT \citep{aly2025ocrapt} detects attacks in provenance graphs with a GNN and has an LLM write an attack report for each subgraph the GNN flags. On DARPA TC3 CADETS we passed all 26 candidate subgraphs, 8 malicious and 18 benign, to the LLM, including benign ones the GNN does not flag. It reported every one as an attack, and still called all 26 malicious after we removed the anomaly-alert wording from the prompt and asked for an explicit malicious-or-benign verdict. The GNN's selection is the only decision source; the report describes that decision without discriminating. RiskTagger \citep{lin2026risktagger} has an LLM grade laundering suspicion for Ethereum addresses. In its released code every graded level maps to label 1, a ``no suspicion'' answer drops the address without a label, and label 0 comes only from a deterministic skip. On 20 addresses attributed to the Bybit hack and 20 presumed-benign addresses, rerunning identical inputs changed the graded level for 21 of 40, injecting upstream high-risk tags changed it for 21 of 40, and the final label changed for at most 1 of 20 in each group. The label comes from the mapping rule, not from the LLM's grading. Both audits used a local Llama 8B model rather than the authors' models, on small samples.

\paragraph{Graph-based AML detection}
Elliptic \citep{weber2019antimoney} and Elliptic++ \citep{elmougy2023demystifying} are the standard Bitcoin benchmarks, and random forests on node features remain strong baselines \citep{weber2019antimoney,maganti2026}; relational transformers have also been applied to card fraud \citep{yan2026temporal}. Distance to known illicit nodes carries signal \citep{oliveira2021guiltywalker}, but proximity reflects graph structure rather than behavior, which limits how well it separates licit from illicit activity \citep{li2025}.

\paragraph{Attribution in multi-agent LLM systems}
The closest line of work is automated failure attribution. \citet{zhang2025whowhen} collect failure logs from 127 LLM multi-agent systems, annotate the responsible agent and the decisive step, and find that LLM-based attributors identify the agent in at most 53.5\% of cases and the step in 14.2\%. Concurrent work uses Shapley-style credit to train or tune agents: \citet{sharp2026} use it as a reward signal, and \citet{hivemind2025} use it for prompt optimization over workflow graphs; teacher uncertainty has also been used to locate the critical decision steps of multi-turn agents for distillation \citep{zhang2026uopd}. In addition, \citet{mu2026counterfactual} define counterfactual responsibility for probabilistic multi-agent systems and relate it to the Shapley value. We differ in three ways. We attribute every decision, including correct ones, not only failures. We compute attribution by intervening on recorded component outputs and re-executing the pipeline, not by asking an LLM to read the log. And we use credit to test whether an evaluation can be trusted, not to optimize the agents.

\paragraph{Causal responsibility, credit, and mediation}
We build on actual causation in the updated Halpern--Pearl definition \citep{halpern2005causes}, which we implement rather than the later modified definition \citep{halpern2015modification}, degree of responsibility \citep{chockler2004responsibility}, and the Shapley value \citep{shapley1953value} in its interventional form \citep{janzing2020feature}. Different Shapley variants can give different answers to the same question \citep{sundararajan2020many}, which is why we fix null policies before computing credit. Shapley credit has been assigned to components inside a network \citep{ghorbani2020neuron}, and natural direct and indirect effects \citep{pearl2001direct} have been estimated from observational data \citep{imai2010general} and by intervention inside language models \citep{vig2020investigating}. In observational settings the cross-world terms of mediation need identification assumptions; in a pipeline that can be re-executed they are computed directly.

\paragraph{Evaluating compound AI systems}
Systems that chain models, retrievers, tools, and rules \citep{zaharia2024compound} are mostly evaluated end to end. \citet{kapoor2025agents} show that accuracy-only agent evaluation has led to mistaken conclusions about where accuracy gains come from, and call for cost-controlled comparisons against simple baselines. Likewise, a valid action and a correct final answer do not show that an agent knows what its action will do \citep{chen2026mongobench}. Our audit is a concrete case of the problem they describe: behavior credited to an LLM judge is produced by a fixed rule. Decision-source attribution adds per-case, per-component evidence to end-to-end metrics.

\paragraph{Validity of LLM judges: anchoring and sycophancy}
LLM-as-a-judge is widely used \citep{zheng2023judging}, and judges show position bias \citep{wang2024fair} and a range of other biases, including deference to authoritative-looking content \citep{ye2025justice}. LLMs are anchored by numeric hints, and chain-of-thought, reflection, and instructions to ignore the anchor do not remove the effect \citep{lou2026anchoring}. A single misleading signal in context flips correct answers across open models, so judges should be assessed on selective trust rather than resistance alone \citep{sun2026scope}. Even when the evidence provably does or does not determine the answer, models answer largely regardless of which \citep{xu2027silica}, and many errors come from how the input is read rather than from reasoning \citep{han2026read}. Models also tend to agree with views stated in the prompt \citep{sharma2024sycophancy}, which parallels automation bias in people who defer to automated aids \citep{parasuraman2010complacency}. Our score-injection and blind-first results fit this picture, and the prompt-level failures reported by \citet{lou2026anchoring} are why we change the pipeline's structure rather than its prompt. These studies test a judge in isolation and ask whether its verdict is correct. In a pipeline, the anchor is often also read by a deterministic rule, so a biased judge and an override can produce the same decision. We ask a prior question, whether the judge's verdict causes the system's decision at all, and only attribution separates the two.

\section{Decision-Source Attribution}

Decision-source attribution asks, for a given case and a given decision, which components of a pipeline produced it. We treat the pipeline as a program that can be re-executed with component outputs replaced, and define three quantities of increasing scope: whether a component was a source of one decision (Layer 1), how much aggregate credit it earns over its own randomness and its interactions with other components (Layer 2), and how much of an upstream signal's effect on the decision passes through it (Layer 3).

\paragraph{Setting}
A pipeline maps a case $x$ to a decision $d \in \mathcal{D}$, a finite set of dispositions (for example $\{\textsc{clear}, \textsc{review}, \textsc{escalate}\}$, or the four dispositions of the audited pipeline: \textsc{sar}, \textsc{rule\_gap}, \textsc{monitor} and \textsc{clear}). It is composed of components $\mathcal{C} = \{c_1,\dots,c_n\}$: models, rules, retrievers, LLM judges and gates. Component $c_i$ produces an output $o_i = f_i(\mathrm{pa}_i, \varepsilon_i)$ from the outputs of its parents $\mathrm{pa}_i$ and its own randomness $\varepsilon_i$ (sampling temperature, retrieval ties); deterministic components have no $\varepsilon_i$. A fixed decision function $g$ maps the outputs to $d = g(o_1,\dots,o_n)$. The trace of a case records every $o_i$. An attribution claim is always about an outcome event $A \subseteq \mathcal{D}$, written $Y_A = \ind[d \in A]$: for example ``flagged'', ``escalated'' or ``seen by a human''. A component can be the source of one event and not another, so every claim below names its event.

\paragraph{Interventions and null policies}
For each component we fix in advance a null policy $\nu_i$: the output the rest of the pipeline receives when $c_i$ is removed. It must be a valid value of $c_i$'s interface, so the pipeline still runs, and it encodes what the pipeline would do without the component: an LLM judge that is off abstains, a verification gate that is off passes its input through, a rule that is off never fires. The intervention $\mathrm{do}(c_i{=}\nu_i)$ replaces $o_i$ with $\nu_i$ and leaves every other mechanism unchanged. Null policies are part of the claim, so we state them with each result and report how sensitive the credit is to the alternatives.

\paragraph{Layer 1: decision source of a single case}
Component $c_i$ is a \emph{but-for decision source} of event $A$ on case $x$ if
\[
Y_A(x) \neq Y_A\bigl(x;\, \mathrm{do}(c_i{=}\nu_i)\bigr),
\]
with every other component at the value it took in the trace. When several components could each produce the decision alone, no single one is a but-for source, so we use actual causation in the updated Halpern--Pearl form \citep{halpern2005causes}. A contingency $W \subseteq \mathcal{C} \setminus \{c_i\}$, set to non-actual values $w'$, is valid if (AC2a) some alternative value of $c_i$ changes the event under $W{=}w'$, and (AC2b) with $c_i$ at its actual value the event keeps its actual value for every subset of $W$ set to $w'$ and every subset of the intermediate variables held at their actual values. The degree of responsibility of $c_i$ is $1/(1+k)$, where $k$ is the size of the smallest valid contingency, and $0$ if none exists \citep{chockler2004responsibility}; but-for is the case $k=0$. Causes are single components, so minimality holds trivially, and contingencies range over components only, with intermediates entering through the AC2b clause. When the trace does not record a component's output (our gate result is stored only when it fires), we compute responsibility under every value consistent with the record and report the range. Layer 1 answers the auditor's question for one case and one realized trace.

\paragraph{Layer 2: interventional credit}
A single trace hides two things: a stochastic component may decide differently on a rerun, and components interact (a judge's flag matters only if a gate lets it through). We therefore define, for each case and event, the value of a coalition $S \subseteq \mathcal{C}$ as the probability of the event when the components in $S$ run and the others are at their nulls,
\[
v_x(S) = \mathbb{E}_{\varepsilon}\Bigl[Y_A\bigl(x;\, \mathrm{do}(c_j{=}\nu_j \text{ for } j \notin S)\bigr)\Bigr],
\]
and the credit of $c_i$ as its interventional Shapley value \citep{janzing2020feature},
\[
\phi_i(x) = \sum_{S \subseteq \mathcal{C}\setminus\{c_i\}} w_S \bigl[v_x(S \cup \{c_i\}) - v_x(S)\bigr],
\]
with $w_S = |S|!\,(n-|S|-1)!/n!$. The credits add up to $v_x(\mathcal{C}) - v_x(\emptyset)$, the change in the event's probability between the full pipeline and the all-null pipeline. Credit is signed: a component that moves the decision away from $A$, such as a gate that cancels a judge's flags, earns negative credit. Group credit is the mean of $\phi_i(x)$ over the cases of a group. We report credit by group (licit controls, illicit cases, score bands), never only overall, because a component can add false alarms in one group and catches in another.

\paragraph{Layer 3: mediated influence of upstream signals}
Layers 1 and 2 say which component decides, not why. When an upstream signal $s$ (a detector's risk score) is read both by the judge and by the deterministic policy, part of its effect passes through the judge and part bypasses it. Writing $Y(s_R, J(s_J))$ for the outcome when the policy reads score $s_R$ and the judge reads score $s_J$, we use the natural direct and indirect effects \citep{pearl2001direct} of moving $s$ from $s$ to $s'$:
\begin{align*}
\mathrm{NDE} &= \mathbb{E}\bigl[Y(s', J(s))\bigr] - \mathbb{E}\bigl[Y(s, J(s))\bigr],\\
\mathrm{NIE} &= \mathbb{E}\bigl[Y(s, J(s'))\bigr] - \mathbb{E}\bigl[Y(s, J(s))\bigr],
\end{align*}
with total effect $\mathrm{TE} = \mathbb{E}[Y(s', J(s'))] - \mathbb{E}[Y(s, J(s))]$. NIE measures how much the judge follows the score (anchoring); NDE measures how much the rules act on it regardless of the judge. Because the pipeline is a program, the cross-world terms are computed directly, by sampling the judge under one score and running the policy under another, so no identification assumptions are needed. With a nonlinear decision function, TE need not equal NDE + NIE, so we report all three.

\paragraph{Estimation}
Algorithm~\ref{alg:dsa} gives the estimator. Upstream evidence (the investigation) is gathered once per case and frozen, so every coalition and every score condition is judged on byte-identical evidence; credit is therefore conditional on that investigation. Because the investigator saw the scores, and one of its tools returned label-derived fields in 17 of 154 investigations, $e_x$ can itself encode the score and, in those cases, the label; Layers 2 and 3 measure credit given this evidence, not over a score-free investigation. Stochastic components are sampled $K$ times per case, and components that consume another stochastic output (a gate checking a judge's citations) are sampled $M$ times per upstream sample, which gives an unbiased nested estimate of each $v_x(S)$. With $n$ small (three or four players), all $2^n$ coalitions are evaluated exactly. Confidence intervals come from a hierarchical bootstrap that resamples cases and, within each case, the stochastic samples.

\begin{algorithm}[tb]
\caption{Decision-source attribution (Layers 2 and 3)}
\label{alg:dsa}
\textbf{Input}: cases $X$; components $\mathcal{C}$ with null policies $\nu$; event $A$; samples $K, M$; score levels $\Sigma$ (Layer 3)
\begin{algorithmic}[1]
\FOR{each case $x \in X$}
\STATE run upstream stages once; freeze the evidence $e_x$
\FOR{$k = 1,\dots,K$}
\STATE sample each stochastic component on $e_x$ (seed $k$, shared across conditions; once per score level in Layer 3)
\STATE sample each nested component $M$ times on that sample
\ENDFOR
\FOR{each coalition $S \subseteq \mathcal{C}$}
\STATE $\hat v_x(S) \gets$ mean of $Y_A$ over samples, with $c_j \gets \nu_j$ for $j \notin S$
\ENDFOR
\STATE $\hat\phi_i(x) \gets$ Shapley value of $c_i$ in the game $\hat v_x$
\STATE Layer 3: for each $(s, s') \in \Sigma^2$, estimate NDE, NIE, TE by running the policy on $s_R$ with judge samples drawn under $s_J$
\ENDFOR
\STATE report group means of $\hat\phi_i$, NDE, NIE, TE with hierarchical bootstrap CIs
\end{algorithmic}
\end{algorithm}

\paragraph{When attribution is deductive, and when it is not}
Some attribution results follow from the pipeline's design and one observed fact, and carry no further information. If the decision is a monotone disjunction $Y = J \lor O$ of a judge flag $J$ and an override $O$, and the judge never flags on the cases studied, then $J$ is a null player: $\phi_J = 0$ and $O$ receives all the credit in every case, by the Shapley null-player property. This is the situation of the audited pipeline's initial runs, and it is why we report its headline result as a consequence of the verdict distribution, not as a separate finding. Attribution becomes an empirical question when (i) components are stochastic, so a component's contribution depends on how often it acts; (ii) the decision function is non-monotone, so a component can push decisions in either direction and its signed credit cannot be read off the design; (iii) components interact, for example when escalation requires both a supported judge flag and a firing rule, so credit depends on their joint behavior; or (iv) an upstream signal reaches the decision by several paths. These are exactly the cases the three layers are built for, and our worked examples cover each: (i) the full-prompt Layer 2 run, (ii) a two-sided override, (iii) the blind-first structure, and (iv) Layer 3.

\paragraph{Structural properties}
Some properties follow from a pipeline's structure for any data. We state them as propositions before reporting numbers, so that they are not mistaken for findings, and the attribution computation checks each one on every case.

\begin{proposition}[null judge]
If the event is a disjunction $Y = J \lor O$ of a judge flag $J$ and an override $O$, and $J$ never flags on the cases studied, then $\phi_J = 0$ and $\phi_O = v(\mathcal{C}) - v(\emptyset)$ in every case.
\end{proposition}
\begin{proof}
$J$'s marginal contribution is zero in every coalition, so it is a null player; efficiency gives the rest.
\end{proof}

\begin{proposition}[blind-first]
Let escalation require a gate-supported judge flag and a firing rule $W$, with the judge's null an abstention and the gate's null pass-through. Then (a) $\phi_J = \phi_W$ for escalation; (b) $\phi_{GE} = 0$ for the event ``seen by a human''; (c) if $W$ does not fire on a case, $\phi_W = 0$ for every event on that case.
\end{proposition}
\begin{proof}
(a) $v(S) = 0$ unless both $J$ and $W$ are in $S$, so swapping them leaves $v$ unchanged and the symmetry axiom applies. (b) A supported flag ends in escalate or review and an unsupported one in review, so the gate never changes whether a human sees the case: it is a null player for that event. (c) $W$'s actual value equals its null, so $v(S \cup \{W\}) = v(S)$.
\end{proof}

\begin{proposition}[no direct path to the flag]
In our current structure the rules read the score only through the cut $\textsc{high} = [s \ge 0.70]$, and the final flag does not read $\textsc{high}$. Hence for the flag $\mathrm{NDE} = 0$ and $\mathrm{TE} = \mathrm{NIE}$ at every pair of score levels; for escalation to SAR, $\mathrm{NIE}(0.1 \to 0.9) = 0$ and $\mathrm{TE} \neq \mathrm{NDE} + \mathrm{NIE}$ whenever the judge's rule-match rate differs between the two levels.
\end{proposition}
\begin{proof}
The first part holds because the flag depends on $s$ only through the judge's prompt. For SAR, $D_{\mathrm{sar}} = \textsc{high}(s_R)\, m(s_J)$ with $m$ the judge's rule-match rate; substituting into the definitions gives $\mathrm{TE} = m(0.9)$, $\mathrm{NDE} = m(0.1)$ and $\mathrm{NIE} = 0$.
\end{proof}

\paragraph{Matched negative controls}
Each attribution claim is paired with a control that holds everything fixed except one input and states in advance what result would falsify the claim (Table~\ref{tab:controls}). Three controls hold by construction (marked S): they check the implementation, not the data, and we report them only as such.

\begin{table*}[t]
\centering
\small
\begin{tabular}{p{0.05\textwidth}p{0.22\textwidth}p{0.19\textwidth}p{0.19\textwidth}p{0.2\textwidth}}
\toprule
Layer & Control & Held fixed & Varied & Falsified if \\
\midrule
L1 & Irrelevant variable (synthetic) & outcome $A \land B$ & an unread variable $Z$ & $\mathrm{resp}(Z) \neq 0$ \\
L1 & No path to the flag (S) & 160 traces, SCM & citation flag, score band & responsibility $> 0$ for the flag \\
L2 & Irrelevant player (synthetic) & outcome $A \land B$ & an unread player's samples & $\phi \neq 0$ or its CI $\neq [0,0]$ \\
L2 & Rule that does not fire (S) & evidence, samples & rule on/off & $\phi_{\mathrm{rule}} \neq 0$ \\
L2 & Same-seed repeat & full request & only when it is sent & the reply differs \\
2$\times$2 & Label permutation, evidence shape & 14 shape features & band labels (1,000 perm.) & permuted AUC far from 0.5 \\
2$\times$2 & Shape features ignore scores & one record & planted scores, flipped rule & any feature changes \\
2$\times$2 & No score in blind prompts & 120 records & every score, every spelling & any hit outside clause text \\
L3 & Blind sanity cell & blind prompts, seeds, weights & injected score (0.1 vs 0.9) & NIE beyond the measured noise floor \\
L3 & Injection never reaches blind prompt & frozen record & all score levels & any blind prompt changes \\
L3 & Gate independent of the score & evidence, clauses & all 7 levels & any gate prompt changes \\
L3 & Real-score leak scan & all but the level & injected and hidden levels & real score outside clause text \\
L3 & Paired-seed estimator null (S) & identical samples & bootstrap resamples & any replicate's NIE $\neq 0$ \\
\bottomrule
\end{tabular}
\caption{Matched negative controls. S: holds by construction (checks the implementation). The artifact and test implementing each control, with its result, are available from the authors on request.}
\label{tab:controls}
\end{table*}

\paragraph{Validation on synthetic pipelines}
Every estimator is tested against answers derived by hand, in exact fractions, before it is applied to the pipeline. For Layer 1, the engine recovers the textbook cases: in disjunctive overdetermination each sufficient cause has responsibility 1/2; in an 11--0 vote each voter has 1/6; an unread variable has 0; and on a non-monotone gate model it matches hand-derived contingencies where a reading without the AC2b clause gives the wrong answer. For Layer 2, the Shapley engine reproduces closed forms for conjunctive, disjunctive and signed gate--override games, and the sampled and nested estimators are unbiased. For Layer 3, the estimator recovers direct-only, indirect-only and interacting pipelines (where $\mathrm{TE} \neq \mathrm{NDE} + \mathrm{NIE}$).

\section{Experimental Setup}

\paragraph{Pipeline}
The pipeline is fully local (Figure~\ref{fig:pipeline}). A temporal transaction detector (Balanced-BiEGCN \citep{xiao2025balanced}) scores each transaction, and a wallet-level graph model (WalletGAT) scores the wallets linked to it. WalletGAT is a two-layer GATv2 network (64 hidden units, 4 attention heads in the first layer) over the Elliptic++ wallet graph (822,942 wallets, 5.7M wallet-to-wallet edges, 169 time-aggregated features per wallet), trained with class-weighted cross-entropy and neighbor sampling; a 5-seed ensemble reaches test AUC-ROC 0.70 and illicit-class F1 0.28. It is trained on a temporal split of the labelled wallets (last activity at timesteps 1--34 for training, 35--41 for validation and 42--49 for test), and none of the 37,820 transactions at timesteps 35--42 links directly to a training wallet. A ReAct investigator gathers transaction and neighborhood evidence with a bounded tool budget. A compliance agent retrieves regulatory clauses from a corpus of six regulatory documents with hybrid FAISS and BM25 retrieval and cross-encoder reranking; citations that cannot be matched to retrieved text are stripped, and the agent may abstain. An LLM judge (Llama 3.1 8B unless stated) issues a verdict, and an entailment gate downgrades a flag when the cited clauses do not support it. A deterministic wallet-score override then upgrades the verdict to a flag when a linked wallet scores at least 0.90, a rule anchored to FATF Recommendation 20. Escalation to a suspicious activity report additionally requires a transaction score of at least 0.70.

\paragraph{Pipeline structures}
The \emph{current} structure is the one above. The \emph{blind-first} structure hides every score and label-leaking field from the judge's prompt, runs the wallet rule separately, and routes cases where the judge and the rule disagree, or where the gate does not support a flag, to human review instead of resolving them automatically.

\paragraph{Evaluation set}
All attribution runs use the same 120 cases: 40 wallet-linked hard cases and 80 licit controls (40 \textsc{high}\_licit and 40 \textsc{low}\_licit by transaction score). The hard cases count as illicit under the wallet-linked label, but at the transaction level 8 are illicit and 32 are licit, so we report the 8 transaction-illicit cases separately. Throughout, \emph{hard case} means wallet-linked hard case, and \emph{illicit} without qualification means illicit at the transaction level. The wallet-linked label is not independent of the transaction label: Elliptic++ labels a wallet illicit if it takes part in any illicit transaction \citep{elmougy2023demystifying}, so for an illicit transaction the hard-case condition holds by construction. The 80 licit controls are transaction-licit, scored transactions before the timestep-43 shift, drawn 40 per band at a transaction-score cut of 0.5 with a fixed seed. Because WalletGAT's training window overlaps them, we also drew 80 out-of-sample controls by the same rule, restricted to timesteps 35--42 and to transactions with no wallet within two wallet-graph hops (WalletGAT's receptive field) active at timesteps 1--34.

\paragraph{Datasets}
Elliptic++ \citep{elmougy2023demystifying}: 203K Bitcoin transactions and 822K wallets over 49 timesteps, with illicit, licit and unknown labels on anonymous features. It is the main testbed for every layer.

IBM AMLworld HI-Small \citep{altman2023realistic}: synthetic bank transfers with ground-truth laundering labels from a simulator that plants known typologies (fan-in, fan-out, cycles, scatter-gather and others). Unlike Elliptic++, it has real amounts, currencies and payment formats, and laundering is rare: 956 of 862,792 transfers (0.11\%) on the test days. We use it to test whether the findings transfer beyond Bitcoin; we report flags only and do not run the three attribution layers on it. First, there is no trained wallet model for AMLworld, and training one on the same labels would make the override circular, so the override reads a label-free proxy: for each account, the highest detector score among its other transfers on the test days, with the transaction under investigation left out. Second, its threshold is the 99th percentile of the proxy scores ($\theta = 0.774$), fixed before any result was recorded. Third, the last days of the simulation (days 11 to 18) are 59\% laundering (655 of 1,108 transfers), so they are excluded from every reported test metric.

\paragraph{Compute}
LLM components ran under Ollama on a cluster node with an NVIDIA Tesla V100 (32 GB). The context length was set explicitly to 32k tokens after we found that leaving it unset lets the server choose it from available memory; a 4k run is a controlled truncation condition, not an alternative configuration. Layer 2 sampled the judge $K=5$ times per case and the gate on each judge sample, with seeds shared across coalitions; the 2$\times$2 used $K=5$, $M=2$ and compared Llama 3.1 8B with Qwen2.5 14B-Instruct. The judge and the gate were sampled at temperature 0.2 and the investigator's controller at 0.1, with JSON output. Every call carried a seed derived from a hash of the case, the component and the sample index, and transport retries reused it. The judge-only grids (score injection, context length, clause counterfactual) ran under Ollama 0.32.1 on an Apple M3 Pro (Llama 3.1 8B, Qwen2.5 7B and 14B) and an NVIDIA H200 (Llama 3.1 70B), all Q4\_K\_M, at temperature 0.5 with 5 seeds, a context of 8,192 tokens and at most 2,048 output tokens; the same Llama 8B weights on both machines gave the same flag on 97.5\% of 479 paired calls. The out-of-sample runs used a laptop CPU, Ollama 0.40.3 and the pipeline's temperatures, with one sample per case and no fixed seed; between their 4k and 32k runs at most 2 of 74 cases and 1 of 80 controls changed their flag.

\section{Results}

Each result is tagged as a \emph{main finding}, a \emph{design consequence} (it follows from the design or the layer definitions, not from model behavior) or a \emph{transfer test}.

\paragraph{Detection baselines}
Table~\ref{tab:detect} shows that the graph detector is not the strongest model by F1 on either dataset. On AMLworld the detectors are weak in absolute terms at this imbalance and each wins on one metric, so we report both.

\begin{table}[tb]
\centering
\begin{tabular}{llcc}
\toprule
Dataset & Model & F1 & AUC-PR \\
\midrule
Elliptic++ & Random forest & 0.61 & -- \\
Elliptic++ & BiEGCN & 0.56 & -- \\
AMLworld & Random forest & 0.075 & 0.040 \\
AMLworld & BiEGCN ensemble & 0.015 & 0.065 \\
\bottomrule
\end{tabular}
\caption{Detection of the illicit class on the test period (AMLworld: days 9--10, tail excluded; RF is the mean of 5 seeds, BiEGCN the 5-seed ensemble without recalibration, which lowers its F1 on AMLworld). On Elliptic++ (test timesteps 42--49), BiEGCN is trained on timesteps 1--39, folding the validation-period labels into training, and is a 5-seed ensemble recalibrated per test timestep with that timestep's illicit rate (aggregate test-period information, no individual labels), and reaches AUC-ROC 0.87; RF uses no recalibration.}
\label{tab:detect}
\end{table}

\subsection{RQ1: Which component decides?}

\paragraph{A silent failure: context truncation [main finding]}
The initial runs never set the judge's context length, so the server silently truncated prompts that exceeded the window it chose. In the Layer 2 rerun on the same 120 cases, the judge returned a verdict in all 120 with a 32k context, and no verdict in 70\% with a 4k context. In the original runs the judge never flagged a hard case, so by Proposition 1 the override received all the credit by construction. That is a consequence of the truncation and the design, and we do not report it as a finding about the judge. A controlled version on the judge alone separates the two failures. With the context capped at 2,048 tokens, which truncates the system prompt, Llama 3.1 8B and Qwen2.5 14B each returned no valid verdict in any of 720 calls on 24 held-out Elliptic++ transactions, whatever score was shown. At 8,192 tokens it copied the score (Table~\ref{tab:inject}), and with the score line removed it flagged 89\% of illicit and 87\% of licit transactions (Qwen2.5 14B: 33\% and 30\%). Truncation, a visible score and a hidden score thus give three different failures, and in none of them does the evidence decide.

\paragraph{With the full prompt, the override still carries the flags [main finding]}
Table~\ref{tab:shapley} gives Layer 2 credit for the event ``flagged'' with the full 32k prompt. The pipeline flags a case with probability 0.587 on average. The wallet override earns $+0.554$, about 94\% of that total. The judge earns $+0.119$, but only on licit controls, where every flag is a false alarm, and the gate earns $-0.087$ because it cancels most of those flags. On the 40 hard cases the override takes all the credit (0.775) and the judge and gate take exactly 0, because the judge never flags a hard case; since the wallet rule fires on only 4 of the 8 transaction-illicit cases, most of that credit is for flagging transactions that are licit at the transaction level. The judge therefore acts once it sees its full prompt, but its actions are false alarms that the gate removes. Over all 120 cases the intervals are $[0.47, 0.64]$ for the override, $[0.09, 0.15]$ for the judge and $[-0.12, -0.06]$ for the gate (95\%, cases resampled; resampling the LLM samples as well changes them by at most 0.001). With the prompt truncated to 4k tokens the override still carries the flags ($+0.497$, $[0.42, 0.57]$), but the judge's credit is indistinguishable from zero ($-0.012$, $[-0.047, 0.028]$) and the gate's turns positive ($+0.095$, $[0.050, 0.137]$), so truncation changes which component appears to act, not only how often. Layer~1 gives the same picture on the 160 stored traces of the original runs: for the flag outcome the override is a but-for cause in 159 of 160, and on the 70 flagged cases it is a but-for cause in all 70 while the judge is one in at most 6.

\begin{table}[tb]
\centering
\begin{tabular}{lccc}
\toprule
Group & Override & Judge & Gate \\
\midrule
All 120 & $+0.554$ & $+0.119$ & $-0.087$ \\
Licit controls (80) & $+0.444$ & $+0.179$ & $-0.130$ \\
Hard cases (40) & $+0.775$ & $0$ & $0$ \\
\bottomrule
\end{tabular}
\caption{Layer 2 interventional Shapley credit for the event ``flagged'', current structure, Llama 3.1 8B, 32k context. Credits sum to the change in flag probability from the all-null pipeline.}
\label{tab:shapley}
\end{table}

\paragraph{The gate decides on clause composition [main finding]}
On the 160 real-pipeline assessments of the disagreement sample, the entailment gate fired on 2/160 cases (1.25\%, 95\% CI [0.34\%, 4.44\%]). That rate comes from the original runs, where truncation left the judge with few flags to check; with the full prompt the gate cancels most of the judge's flags (Table~\ref{tab:shapley}). The gate prompt contains no behavioral features, and all 20 forced-conflict cases shared one evidence bundle, so we read this as the gate responding to which clauses were retrieved, not to the activity it never sees. Citation verification behaves as designed: 37 of 160 assessments (23.1\%) had at least one ungrounded citation stripped, and all 202 surviving citations matched retrieved text.

\paragraph{A non-monotone override: the layers disagree on preemption [design consequence]}
To test attribution where the answer cannot be read off the design, we replace the one-sided override by a two-sided one: it still raises a clear verdict to a flag when a linked wallet scores at least 0.90, and it now clears a flag that passed the gate when every linked wallet scores below $\tau = 0.50$, the wallet model's own decision threshold. The variant is evaluated counterfactually on the stored judge and gate samples of both backbones, which is valid because neither model sees the rule; at $\tau = 0$ it reproduces the Layer 2 results exactly. The threshold was first set to 0.10 and moved to 0.50 before any credit was computed, once the wallet scores showed that at 0.10 no case the judge flags could be affected; 0.05, 0.10 and 0.20 are reported as sensitivity. The downgrade changes the outcome on one case (Llama; flag probability 0.20 to 0) and on none for Qwen. Yet it earns Shapley credit of $-1/6$ on eight case--backbone pairs where the gate already rejects every flag, because Shapley averages over coalitions in which the gate is switched off and the downgrade does the clearing. On the same cases, actual causation gives the downgrade zero responsibility, since in the realized trace the gate fired first and preempted it, and halves the gate's responsibility (from 1 to 1/2). Both answers are correct for their question: Layer 2 measures what a component contributes across configurations, Layer 1 what it did in this trace. The disagreement is why we report both.

\subsection{RQ2: Evidence or score?}

\paragraph{The judge follows the score, not the evidence [main finding]}
We tested score injection on the judge alone, with the gate and the override off, over 76 distinct evidence bundles in three cohorts: 24 known-answer cases on a synthetic policy; 28 matched violating and compliant cases over 14 AML typologies, identical except for their decisive facts and citing verbatim FATF, EU and FinCEN clauses; and 24 held-out Elliptic++ transactions from timestep 42 with their pipeline clauses. Each case was judged at injected scores 0.10 and 0.90 on byte-identical evidence, by four backbones, under the pipeline's prompt, which tells the judge to match its verdict to a score-derived risk tier, and under a neutral prompt that calls the score an uncertain estimate. Table~\ref{tab:inject} gives the score effect, the share of paired verdicts that flip when only the score changes. The deployed configuration, Llama 3.1 8B with the pipeline prompt, follows the score on nearly every case in every cohort. Scale and prompt help only where the evidence is readable: on the typology cases, Llama 3.1 70B's effect falls to +15\%, and the neutral prompt brings every backbone except Llama 8B to +5--10\%. On the real Elliptic++ transactions all four backbones follow the score on every pair under the pipeline prompt, and Llama 70B does so under the neutral prompt too. On the typology forced-conflict arms, where the evidence is correct by construction (a violating case shown 0.10, a compliant one shown 0.90), the deployed configuration follows the wrong score in 96\% of calls, against 16\% for Llama 70B and 42\% for Qwen 14B. The judge can decide from readable evidence, but on the anonymous features the pipeline gives it, the score decides. Deferring to a detector with AUC-ROC 0.87 is not irrational when the features are unreadable; the forced-conflict arms show the deployed judge defers even when readable evidence contradicts the score.

\begin{table}[tb]
\centering
\small
\setlength{\tabcolsep}{1mm}
\begin{tabular}{lcccccc}
\toprule
 & \multicolumn{2}{c}{Synth.} & \multicolumn{2}{c}{Typology} & \multicolumn{2}{c}{Elliptic++} \\
\cmidrule(lr){2-3}\cmidrule(lr){4-5}\cmidrule(lr){6-7}
Model & P & N & P & N & P & N \\
\midrule
Llama 8B & +100 & +43 & +96 & +48 & +100 & +57 \\
Llama 70B & 0 & +12 & +15 & +5 & +100 & +100 \\
Qwen 7B & +81 & +15 & +86 & +9 & +100 & +13 \\
Qwen 14B & 0 & 0 & +42 & +10 & +100 & 0 \\
\bottomrule
\end{tabular}
\caption{Score effect of the judge alone (percent): flag rate at injected score 0.90 minus flag rate at 0.10 on byte-identical evidence, averaged over 5 seeds within case and then over cases. +100 means the score alone decides. P: pipeline prompt; N: neutral prompt. Qwen2.5 14B's 0 on Elliptic++ under the neutral prompt is because it never flags. Case-bootstrap 95\% intervals are available from the authors on request.}
\label{tab:inject}
\end{table}

\paragraph{Verified citations need not apply [main finding]}
To test whether the verdict depends on the clauses, we kept each typology case's facts fixed and replaced its clauses with another typology's clauses that do not apply to it. With the score hidden, the deployed configuration still flags 90\% of violating cases under inapplicable clauses (100\% with the matched ones), and Llama 3.1 70B flags 97--100\% under either prompt; only Qwen2.5 14B with the neutral prompt reads the clauses (99\% to 41\%, with compliant cases near 0\%). A substantial share of flags under inapplicable clauses quote one of them verbatim as the basis (Llama 8B 50 of 123, Llama 70B 103 of 138, Qwen 14B 87 of 98), and every such quote passes the pipeline's substring check. A verified citation shows that the text exists, not that it applies, just as grounding an answer to its location in a document \citep{wen2026switching} shows where it came from, not that it is correct; scale does not make the judge check.

\paragraph{Layer 3: the judge copies the score, and the pipeline absorbs it [main finding]}
We swept the score shown to the judge over 0.1, 0.3, 0.5, 0.7 and 0.9 and a hidden condition on all 120 cases, with the evidence frozen and $K=5$ judge samples per level, and crossed it with the score the rules read. Llama's verdict is a step function of the score: it flags no sample at 0.1 or 0.3, 15\% at 0.5, and every sample at 0.7, at 0.9 and when the score is hidden. Qwen rises more gradually (0\%, 0\%, 16\%, 45\% and 62\%; 58\% hidden). Yet the final flag moves by only 0.037 [0.013, 0.063] for Llama and by exactly 0 for Qwen: the gate rejects almost every judge flag (it passed 32 of 600 for Llama and none for Qwen), and the wallet override decides the rest. By Proposition~3 the flag has no direct path, so this small effect is entirely indirect. For escalation to a SAR the score acts mainly through the rules' 0.70 cut: the direct effect is 0.32 [0.25, 0.39] for Llama and 0.55 [0.46, 0.64] for Qwen, the indirect effect is zero by construction, and the interaction ($+0.09$ and $-0.07$) enters through which clauses the judge cites, not through its verdict. In the blind-first sanity cell, where nothing reads the score, the indirect effect is 0, and 600 of 600 judge replies are byte-identical across the two injected scores. The judge copies the score, but the gate and the override, not the judge, carry it to the decision.

\subsection{RQ3: Do the answers hold?}

\paragraph{Structure matters more than the model [main finding]}
Table~\ref{tab:2x2} reports the 2$\times$2 on the same frozen evidence. In the current structure the two backbones behave alike. With the score visible, Llama flags 86\% of \textsc{high}\_licit samples and 0\% of \textsc{low}\_licit ones; Qwen flags 57\% and 0\%. Both answer ``no clear violation'' in all samples on the 8 transaction-illicit cases, so only the 4 of those on which the wallet rule fires can reach a human. In the blind-first structure Llama flags every case, which shows that its earlier verdicts were reading the score. Qwen flags 72\% of \textsc{high}\_licit, 37\% of \textsc{low}\_licit and 75\% of transaction-illicit samples, clears 42\% of \textsc{low}\_licit controls, and sends all 8 transaction-illicit cases to a human.

That is weaker evidence of judgment than it looks. Qwen flags \textsc{high}\_licit more than \textsc{low}\_licit cases although both are licit and the score is hidden. The blind evidence still predicts \textsc{high} versus \textsc{low} band with out-of-fold AUC 0.80 against a permuted 0.49 ($p = 0.001$), because the investigator that gathered it saw the score: its strongest shape features are the number of wallet lookups, tool errors, and whether it hit the step limit. Retrieval was also tiered by score: on the typology cases its queries recover 14\% of the case-specific clauses, against 52\% for queries built from the case facts, and return the same clauses for a violating case and its control. On hard cases Qwen flags transaction-illicit cases only slightly more than transaction-licit ones (75\% vs.\ 62\%, $n=8$, not significant). Changing the structure changes outcomes a lot for both models; changing the model matters little in the current structure and a lot in blind-first. By Proposition 2, the judge and the wallet rule share escalation credit equally in blind-first, and the gate's credit for ``seen by a human'' is zero, so we do not report those numbers as findings.

\begin{table}[tb]
\centering
\small
\setlength{\tabcolsep}{1mm}
\begin{tabular}{lcccc}
\toprule
 & \multicolumn{2}{c}{Judge flag rate} & \multicolumn{2}{c}{Seen by a human} \\
 & H/L licit & tx-illicit & licit (80) & tx-illicit (8) \\
\midrule
Llama, current & 86 / 0 & 0 & 50 & 35 \\
Qwen, current & 57 / 0 & 0 & 57 & 45 \\
Llama, blind & 100 / 100 & 100 & 100 & 100 \\
Qwen, blind & 72 / 37 & 75 & 73 & 100 \\
\bottomrule
\end{tabular}
\caption{Backbone $\times$ structure on 120 cases (percent of samples). ``Seen by a human'' is escalate or review; on licit controls every such case is a false alarm. H/L: \textsc{high}\_licit / \textsc{low}\_licit.}
\label{tab:2x2}
\end{table}

\paragraph{Out of sample, the pipeline does not separate illicit from licit cases [main finding]}
To measure discrimination, we built an out-of-sample hard set: transactions at timesteps 35--42 that the detector clears (score $<0.5$), under the out-of-sample wallet restriction above, crossed by transaction label and by whether WalletGAT scores any linked wallet at 0.5 or above (26 illicit and 48 licit cases; the 6 illicit cases WalletGAT misses are all included). With the full 32k prompt the pipeline flags 12 of 26 illicit (46.2\%) and 18 of 48 licit (37.5\%) cases, LR$^{+}$ 1.23 (95\% CI 0.71--2.14); with the context truncated to 4k, 10/26 and 18/48 (LR$^{+}$ 1.03). Every flag comes from the wallet override; the judge flags none of the 74 on its own, and no override threshold between 0.50 and 0.99 gives LR$^{+}$ above 1.2. On the 80 out-of-sample licit controls the pipeline flags 37 (46.3\%, 35.7--57.1\%), against 39 of 80 (48.8\%) in sample, so the false alarms are not an artifact of WalletGAT's training overlap; 36 of the 37 come from the override.
Neither the controller nor the score changes this. With the LLM controller replaced by the fixed safeguard policy, the 32k result is identical (12/26, 18/48). Hiding the score from the judge leaves it unchanged (12/26, 19/48, LR$^{+}$ 1.17), but only because the gate cancels 70 of the judge's 74 flags. The signal is in the graph: a logistic combiner over ten label-free graph features, trained on timesteps 1--34, reaches LR$^{+}$ 3.19 (2.34--4.36; 38.3\% recall at 12.0\% false-positive rate) on all 81 illicit and 1,352 licit out-of-sample hard cases, against 1.80 for the single-wallet 0.90 rule; it is a proof of concept with 56 illicit training cases.

\paragraph{AMLworld: detectors, override and judge [transfer test]}
A first run returned no verdict because the cluster's HTTP proxy intercepted calls to the local model server; with that fixed, the judge returned a verdict in all 120 assessments. On 20 hard laundering cases, all scoring below 0.5, the pipeline flags 7 (35\%, 95\% CI 18--57\%), 6 of them through the proxy override, against 1 of 40 licit transfers below 0.5 (2.5\%, 0.4--13\%; Fisher $p = 0.001$), also through the override. The margin is thin, because the proxy scores bunch between 0.76 and 0.79. The judge alone flags 1 of the 20 laundering cases, and 3 of 20 when it sees only the transaction. On licit transfers it follows the score band, as on Elliptic++: on its own it flags 13 of 40 above 0.5 and none of 40 below ($p < 0.001$). The override again carries the detection, and the judge adds false alarms on high-score cases.

\section{Discussion}

\paragraph{Explanation is not attribution}
This pipeline produces regulation-citing justifications with verified citations while a single rule carries most of the decision credit. An evaluation that rated explanation quality, as in prior AML systems, would score it well. The two published pipelines we audited (Section~II) show the same pattern: a GNN's selection in OCR-APT and a label-mapping rule in RiskTagger decide, while the LLM writes. Reasoning traces likewise often omit the injected bias that changed an answer \citep{sun2026biasack}.

\paragraph{Accountability across delegation}
In agent pipelines the agent that speaks last is often not the agent that decided. Attribution exposes a form of evaluation gaming that needs no intent: a judge that copies an upstream score passes end-to-end checks while contributing nothing, and hiding the score does not remove the copy if an upstream agent that saw it shaped the evidence. Verification therefore has to follow information through the whole delegation chain.

\paragraph{Design implication}
Blind-first reconciliation makes the judge's contribution measurable, but it is only as blind as its evidence, and it trades the current structure's escalation of high-score licit controls for a much larger review load.

\paragraph{Limitations}
We study one pipeline in depth; the two published pipelines received only targeted interventions. Elliptic++ is real but anonymous, and AMLworld is synthetic, uses a proxy override, excludes the simulation tail, and is evaluated on flags from one sample per case, without the three attribution layers; its typologies are planted by the same simulator that labels them. The attribution set has only 8 transaction-illicit hard cases and the out-of-sample set 26; a larger transaction-illicit sample, ideally the full Elliptic++ test set, with every attribution metric stratified by transaction label, is future work. Several diagnostics use 20 cases, and $K=5$, $M=2$ give wide per-case intervals. Attribution is conditional on frozen evidence gathered by a Llama controller that saw the scores. One investigator tool returned label-derived fields, including the queried wallet's own label, which for an illicit transaction exposes its own label; it was reached in 17 of 154 investigations, and rerunning those without the fields left every licit case unchanged and changed one illicit case, through the override, which never reads those fields. Model family and size change together in the 2$\times$2, and replies are not seed-reproducible across server processes. The two-sided override is a counterfactual variant evaluated on stored samples, not a deployed rule, and its threshold was re-set once before results were computed. In Layer 3, two Qwen levels were resumed after a cluster node killed their jobs, so their samples come from two server processes. The typology cases were written by the authors and reviewed within the team, not adjudicated by compliance officers, and several violating cases paraphrase their clauses, so they test unambiguous evidence rather than legal reasoning.

\section*{Ethical Statement}
Decision-source attribution can help auditors find which component of an automated system made a decision; it could also be used to make a rule-driven system appear judgment-driven, one reason we report structural results separately. All data are public (Elliptic++) or synthetic (AMLworld), and the pipeline's outputs should not be used for enforcement without human review.

\renewcommand{\bibfont}{\footnotesize}
\bibliographystyle{IEEEtranN}
\bibliography{references}

\begin{thebibliography}{42}
\providecommand{\natexlab}[1]{#1}
\providecommand{\url}[1]{#1}
\csname url@samestyle\endcsname
\providecommand{\newblock}{\relax}
\providecommand{\bibinfo}[2]{#2}
\providecommand{\BIBentrySTDinterwordspacing}{\spaceskip=0pt\relax}
\providecommand{\BIBentryALTinterwordstretchfactor}{4}
\providecommand{\BIBentryALTinterwordspacing}{\spaceskip=\fontdimen2\font plus
\BIBentryALTinterwordstretchfactor\fontdimen3\font minus \fontdimen4\font\relax}
\providecommand{\BIBforeignlanguage}[2]{{%
\expandafter\ifx\csname l@#1\endcsname\relax
\typeout{** WARNING: IEEEtranN.bst: No hyphenation pattern has been}%
\typeout{** loaded for the language `#1'. Using the pattern for}%
\typeout{** the default language instead.}%
\else
\language=\csname l@#1\endcsname
\fi
#2}}
\providecommand{\BIBdecl}{\relax}
\BIBdecl

\bibitem[Weber et~al.(2019)Weber, Domeniconi, Chen, Weidele, Bellei, Robinson, and Leiserson]{weber2019antimoney}
M.~Weber, G.~Domeniconi, J.~Chen, D.~K.~I. Weidele, C.~Bellei, T.~Robinson, and C.~E. Leiserson, ``Anti-money laundering in bitcoin: Experimenting with graph convolutional networks for financial forensics,'' 2019, {KDD} Workshop on Anomaly Detection in Finance.

\bibitem[Elmougy and Liu(2023)]{elmougy2023demystifying}
Y.~Elmougy and L.~Liu, ``Demystifying fraudulent transactions and illicit nodes in the bitcoin network for financial forensics,'' in \emph{Proceedings of the 29th ACM SIGKDD Conference on Knowledge Discovery and Data Mining}, 2023, pp. 3979--3990.

\bibitem[Axelsen et~al.(2025)Axelsen, Licht, and Damsgaard]{axelsen2025agentic}
H.~Axelsen, V.~Licht, and J.~Damsgaard, ``Agentic {AI} for financial crime compliance,'' 2025.

\bibitem[Naik et~al.(2025)Naik, Dintakurthi, Hu, Wang, and Qiu]{naik2025coinvestigator}
P.~V. Naik, N.~K. Dintakurthi, Z.~Hu, Y.~Wang, and R.~Qiu, ``Co-investigator {AI}: The rise of agentic {AI} for smarter, trustworthy {AML} compliance narratives,'' 2025.

\bibitem[Khanvilkar and Kommuru(2025)]{khanvilkar2025regulatory}
K.~Khanvilkar and K.~Kommuru, ``Regulatory graphs and {GenAI} for real-time transaction monitoring and compliance explanation in banking,'' 2025.

\bibitem[Aly et~al.(2025)Aly, Mansour, and Youssef]{aly2025ocrapt}
A.~Aly, E.~Mansour, and A.~Youssef, ``{OCR-APT}: Reconstructing {APT} stories from audit logs using subgraph anomaly detection and {LLMs},'' in \emph{Proceedings of the ACM SIGSAC Conference on Computer and Communications Security (CCS)}, 2025.

\bibitem[Lin et~al.(2026)Lin, Ding, Zou, Chen, Luo, Wu, and Zheng]{lin2026risktagger}
D.~Lin, Y.~Ding, W.~Zou, J.~Chen, X.~Luo, J.~Wu, and Z.~Zheng, ``{RiskTagger}: An {LLM}-based agent for automatic annotation of {Web3} crypto money laundering behaviors,'' in \emph{Proceedings of the ACM Web Conference (WWW)}, 2026, arXiv:2510.17848.

\bibitem[Maganti(2026)]{maganti2026}
S.~Maganti, ``When graph structure becomes a liability: A critical re-evaluation of graph neural networks for {Bitcoin} fraud detection under temporal distribution shift,'' 2026.

\bibitem[Yan(2026)]{yan2026temporal}
Q.~Yan, ``Temporal heterogeneous graph transformer for credit card fraud detection,'' \emph{arXiv preprint arXiv:2609.07100}, 2026.

\bibitem[Oliveira et~al.(2021)Oliveira, Torres, Silva, Aparicio, Ascens{\~a}o, and Bizarro]{oliveira2021guiltywalker}
C.~Oliveira, J.~Torres, M.~I. Silva, D.~Aparicio, J.~T. Ascens{\~a}o, and P.~Bizarro, ``{GuiltyWalker}: Distance to illicit nodes in the {Bitcoin} network,'' 2021.

\bibitem[Li et~al.(2025)Li, Jia, and Su]{li2025}
M.~Li, L.~Jia, and X.~Su, ``Global-local graph attention with cyclic pseudo-labels for {Bitcoin} anti-money laundering detection,'' \emph{Scientific Reports}, vol.~15, p. 22668, 2025.

\bibitem[Zhang et~al.(2025)Zhang, Yin, Zhang, Liu, Han, Zhang, Li, Wang, Wang, Chen, and Wu]{zhang2025whowhen}
S.~Zhang, M.~Yin, J.~Zhang, J.~Liu, Z.~Han, J.~Zhang, B.~Li, C.~Wang, H.~Wang, Y.~Chen, and Q.~Wu, ``Which agent causes task failures and when? {On} automated failure attribution of {LLM} multi-agent systems,'' in \emph{Proceedings of the 42nd International Conference on Machine Learning (ICML)}, ser. PMLR, vol. 267, 2025.

\bibitem[Li et~al.(2026)]{sharp2026}
Y.~Li \emph{et~al.}, ``Who deserves the reward? {SHARP}: {Shapley} credit-based optimization for multi-agent system,'' 2026.

\bibitem[Xia et~al.(2026)Xia, Wang, Zhang, Weng, Cao, and Liew]{hivemind2025}
Y.~Xia, T.~Wang, S.~Zhang, Z.~Weng, B.~Cao, and S.~C. Liew, ``{HiveMind}: Contribution-guided online prompt optimization of {LLM} multi-agent systems,'' in \emph{Proceedings of the AAAI Conference on Artificial Intelligence}, vol.~40, no.~35, 2026, pp. 29\,767--29\,774.

\bibitem[Zhang et~al.(2026)Zhang, Xu, Du, Zhang, and Cai]{zhang2026uopd}
W.~Zhang, P.~Xu, W.~Du, J.~Zhang, and H.~Cai, ``{UOPD}: Uncertainty-aware intervention for on-policy distillation of multi-turn agents,'' \emph{arXiv preprint arXiv:2609.34036}, 2026.

\bibitem[Mu and Najib(2026)]{mu2026counterfactual}
C.~Mu and M.~Najib, ``Counterfactual reasoning for responsibility attribution in probabilistic multi-agent systems,'' in \emph{Proceedings of the Thirty-Fifth International Joint Conference on Artificial Intelligence (IJCAI)}, 2026.

\bibitem[Halpern and Pearl(2005)]{halpern2005causes}
J.~Y. Halpern and J.~Pearl, ``Causes and explanations: A structural-model approach. {Part I}: Causes,'' \emph{The British Journal for the Philosophy of Science}, vol.~56, no.~4, pp. 843--887, 2005.

\bibitem[Halpern(2015)]{halpern2015modification}
J.~Y. Halpern, ``A modification of the {Halpern-Pearl} definition of causality,'' in \emph{Proceedings of the Twenty-Fourth International Joint Conference on Artificial Intelligence (IJCAI)}, 2015, pp. 3022--3033.

\bibitem[Chockler and Halpern(2004)]{chockler2004responsibility}
H.~Chockler and J.~Y. Halpern, ``Responsibility and blame: A structural-model approach,'' \emph{Journal of Artificial Intelligence Research}, vol.~22, pp. 93--115, 2004.

\bibitem[Shapley(1953)]{shapley1953value}
L.~S. Shapley, ``A value for n-person games,'' in \emph{Contributions to the Theory of Games {II}}, H.~W. Kuhn and A.~W. Tucker, Eds.\hskip 1em plus 0.5em minus 0.4em\relax Princeton University Press, 1953, pp. 307--317.

\bibitem[Janzing et~al.(2020)Janzing, Minorics, and Bl{\"o}baum]{janzing2020feature}
D.~Janzing, L.~Minorics, and P.~Bl{\"o}baum, ``Feature relevance quantification in explainable {AI}: A causal problem,'' in \emph{Proceedings of the Twenty Third International Conference on Artificial Intelligence and Statistics (AISTATS)}, 2020, pp. 2907--2916.

\bibitem[Sundararajan and Najmi(2020)]{sundararajan2020many}
M.~Sundararajan and A.~Najmi, ``The many {Shapley} values for model explanation,'' in \emph{Proceedings of the 37th International Conference on Machine Learning (ICML)}, ser. PMLR, vol. 119, 2020, pp. 9269--9278.

\bibitem[Ghorbani and Zou(2020)]{ghorbani2020neuron}
A.~Ghorbani and J.~Zou, ``Neuron {Shapley}: Discovering the responsible neurons,'' in \emph{Advances in Neural Information Processing Systems 33 (NeurIPS)}, 2020.

\bibitem[Pearl(2001)]{pearl2001direct}
J.~Pearl, ``Direct and indirect effects,'' in \emph{Proceedings of the Seventeenth Conference on Uncertainty in Artificial Intelligence (UAI)}, 2001, pp. 411--420.

\bibitem[Imai et~al.(2010)Imai, Keele, and Tingley]{imai2010general}
K.~Imai, L.~Keele, and D.~Tingley, ``A general approach to causal mediation analysis,'' \emph{Psychological Methods}, vol.~15, no.~4, pp. 309--334, 2010.

\bibitem[Vig et~al.(2020)Vig, Gehrmann, Belinkov, Qian, Nevo, Singer, and Shieber]{vig2020investigating}
J.~Vig, S.~Gehrmann, Y.~Belinkov, S.~Qian, D.~Nevo, Y.~Singer, and S.~Shieber, ``Investigating gender bias in language models using causal mediation analysis,'' in \emph{Advances in Neural Information Processing Systems 33 (NeurIPS)}, 2020.

\bibitem[Zaharia et~al.(2024)Zaharia, Khattab, Chen, Davis, Miller, Potts, Zou, Carbin, Frankle, Rao, and Ghodsi]{zaharia2024compound}
M.~Zaharia, O.~Khattab, L.~Chen, J.~Q. Davis, H.~Miller, C.~Potts, J.~Zou, M.~Carbin, J.~Frankle, N.~Rao, and A.~Ghodsi, ``The shift from models to compound {AI} systems,'' Berkeley Artificial Intelligence Research Blog, 2024.

\bibitem[Kapoor et~al.(2025)Kapoor, Stroebl, Siegel, Nadgir, and Narayanan]{kapoor2025agents}
S.~Kapoor, B.~Stroebl, Z.~S. Siegel, N.~Nadgir, and A.~Narayanan, ``{AI} agents that matter,'' \emph{Transactions on Machine Learning Research}, 2025.

\bibitem[Chen et~al.(2026)Chen, Qiu, Cheng, and Qian]{chen2026mongobench}
Q.~Chen, H.~Qiu, Y.~Cheng, and M.~Qian, ``{MongoBench}: Do {LLMs} know what a {MongoDB} query will do?'' in \emph{Proc. IEEE CAIT}, 2026, to appear; arXiv:2611.09065.

\bibitem[Zheng et~al.(2023)Zheng, Chiang, Sheng, Zhuang, Wu, Zhuang, Lin, Li, Li, Xing, Zhang, Gonzalez, and Stoica]{zheng2023judging}
L.~Zheng, W.-L. Chiang, Y.~Sheng, S.~Zhuang, Z.~Wu, Y.~Zhuang, Z.~Lin, Z.~Li, D.~Li, E.~P. Xing, H.~Zhang, J.~E. Gonzalez, and I.~Stoica, ``Judging {LLM}-as-a-judge with {MT-Bench} and {Chatbot Arena},'' in \emph{Advances in Neural Information Processing Systems 36, Datasets and Benchmarks Track}, 2023.

\bibitem[Wang et~al.(2024)Wang, Li, Chen, Cai, Zhu, Lin, Cao, Kong, Liu, Liu, and Sui]{wang2024fair}
P.~Wang, L.~Li, L.~Chen, Z.~Cai, D.~Zhu, B.~Lin, Y.~Cao, L.~Kong, Q.~Liu, T.~Liu, and Z.~Sui, ``Large language models are not fair evaluators,'' in \emph{Proceedings of the 62nd Annual Meeting of the Association for Computational Linguistics (ACL)}, 2024.

\bibitem[Ye et~al.(2025)Ye, Wang, Huang, Chen, Zhang, Moniz, Gao, Geyer, Huang, Chen, Chawla, and Zhang]{ye2025justice}
J.~Ye, Y.~Wang, Y.~Huang, D.~Chen, Q.~Zhang, N.~Moniz, T.~Gao, W.~Geyer, C.~Huang, P.-Y. Chen, N.~V. Chawla, and X.~Zhang, ``Justice or prejudice? {Quantifying} biases in {LLM}-as-a-judge,'' in \emph{International Conference on Learning Representations (ICLR)}, 2025.

\bibitem[Lou and Sun(2026)]{lou2026anchoring}
J.~Lou and Y.~Sun, ``Anchoring bias in large language models: An experimental study,'' \emph{Journal of Computational Social Science}, vol.~9, 2026.

\bibitem[Sun et~al.(2026{\natexlab{a}})Sun, Chow, Wang, Liu, Gao, Wu, and Kong]{sun2026scope}
X.~Sun, W.~Chow, Y.~Wang, J.~Liu, W.~Gao, Q.~Wu, and L.~Kong, ``Learning when to trust via selective context preference optimization,'' \emph{arXiv preprint arXiv:2608.06377}, 2026.

\bibitem[Xu and Guan(2027)]{xu2027silica}
P.~Xu and X.~Guan, ``{SILICA}: Certified counterfactual evaluation of identifiability in unseen-language induction,'' in \emph{Findings of EACL}, 2027, to appear.

\bibitem[Han et~al.(2026)Han, Cui, Guo, Wang, and Lyu]{han2026read}
F.~Han, H.~Cui, L.~Guo, Z.~Wang, and Z.~Lyu, ``Read before you think: Mitigating {LLM} comprehension failures with step-by-step reading,'' in \emph{Proc. IEEE ICASSP}, 2026, pp. 17\,107--17\,111.

\bibitem[Sharma et~al.(2024)Sharma, Tong, Korbak, Duvenaud, Askell, Bowman, Cheng, Durmus, Hatfield-Dodds, Johnston, Kravec, Maxwell, McCandlish, Ndousse, Rausch, Schiefer, Yan, Zhang, and Perez]{sharma2024sycophancy}
M.~Sharma, M.~Tong, T.~Korbak, D.~Duvenaud, A.~Askell, S.~R. Bowman, N.~Cheng, E.~Durmus, Z.~Hatfield-Dodds, S.~R. Johnston, S.~Kravec, T.~Maxwell, S.~McCandlish, K.~Ndousse, O.~Rausch, N.~Schiefer, D.~Yan, M.~Zhang, and E.~Perez, ``Towards understanding sycophancy in language models,'' in \emph{International Conference on Learning Representations (ICLR)}, 2024.

\bibitem[Parasuraman and Manzey(2010)]{parasuraman2010complacency}
R.~Parasuraman and D.~H. Manzey, ``Complacency and bias in human use of automation: An attentional integration,'' \emph{Human Factors}, vol.~52, no.~3, pp. 381--410, 2010.

\bibitem[Xiao and Yin(2025)]{xiao2025balanced}
B.~Xiao and W.~Yin, ``{Balanced-BiEGCN}: A bidirectional {EvolveGCN} with a class-balanced learning network for dynamic anomaly detection in {Bitcoin},'' \emph{Entropy}, vol.~27, no.~10, p. 1045, 2025.

\bibitem[Altman et~al.(2023)Altman, Blanu{\v{s}}a, von Niederh{\"a}usern, Egressy, Anghel, and Atasu]{altman2023realistic}
E.~Altman, J.~Blanu{\v{s}}a, L.~von Niederh{\"a}usern, B.~Egressy, A.~Anghel, and K.~Atasu, ``Realistic synthetic financial transactions for anti-money laundering models,'' in \emph{Advances in Neural Information Processing Systems 36, Datasets and Benchmarks Track}, 2023.

\bibitem[Wen et~al.(2026)Wen, Li, Paul, Saidutta, Gunaratna, and Chappidi]{wen2026switching}
X.~Wen, W.~Li, S.~Paul, Y.~M. Saidutta, K.~Gunaratna, and S.~Chappidi, ``Switching heads and softening tokens: Turnkey solutions to visually grounded document {QA},'' in \emph{Findings of ACL}, 2026, pp. 36\,490--36\,503.

\bibitem[Sun et~al.(2026{\natexlab{b}})Sun, Gao, Wang, Kong, Li, Fan, Zhuang, Dong, Zheng, Paranjape, Mandal, and Zhang]{sun2026biasack}
X.~Sun, W.~Gao, Y.~Wang, L.~Kong, Y.~Li, Z.~Fan, Z.~Zhuang, W.~Dong, Z.~Zheng, H.~Paranjape, A.~Mandal, and J.~R. Zhang, ``Beyond accuracy: Measuring bias acknowledgment in chain-of-thought reasoning for responsible {AI} evaluation,'' in \emph{ICML Workshop on AI for Good}, 2026, arXiv:2606.15127.

\end{thebibliography}

\end{document}